# Contrast-agent-induced deterministic component of CT-density in the abdominal aorta during routine angiography: proof of concept study

Maria R. Kodenko[a,b,*], Yuriy A. Vasilev[a], Nicholas S. Kulberg[c], Andrey V. Samorodov[b], Anton V. Vladzimirskyy[a,d], Olga V. Omelyanskaya[a] and Roman V. Reshetnikov[a]

[a]*Research and Practical Clinical Center for Diagnostics and Telemedicine Technologies, Department of Health Care of Moscow, Russian Federation, 127051, Petrovka street, 24, building 1, Mocsow, Russia*

[b]*Bauman Moscow State Technical University, 2nd Baumanskaya street, 5, building 1, 105005, Moscow, Russia*

[c]*Federal Research Center "Computer Science and Control" of Russian Academy of Sciences, Vavilova street, 44, building 2, 119333, Mocsow, Russia*

[d]*Department of Information and Internet Technologies, I.M. Sechenov First Moscow State Medical University (Sechenov University), Trubetskaya Street, 8, Building 2, 119991 Moscow, Russia*

**Abstract**

*Background and objective:* CTA is a gold standard of preoperative diagnosis of abdominal aorta and typically used for geometric-only characteristic extraction. We assume that a model describing the dynamic behavior of the contrast agent in the vessel can be developed from the data of routine CTA studies, allowing the procedure to be investigated and optimized without the need for additional perfusion CT studies. Obtained spatial distribution of CA can be valuable for both increasing the diagnostic value of a particular study and improving the CT data processing tools.

---
[*]Corresponding author
E-mail address: m.r.kodenko@yandex.ru; postal address: 127051, Petrovka street, 24, building 1, Mocsow, Russia

**Abbreviations:**
CA – contrast agent
CAiDC – contrast agent-induced deterministic component
CTA – computed tomographic angiography
HU – Hounsfield unit
NLS – nonlinear least square
ROI – region of interest
RMSE – root mean square error

*Methods:* In accordance with the Beer-Lambert law and the absence of chemical interaction between blood and CA, we postulated the existence of a deterministic CA-induced component in the CT signal density. The proposed model, having a double-sigmoid structure, contains six coefficients relevant to the properties of hemodynamics. To validate the model, expert segmentation was performed using the 3D Slicer application for the CTA data obtained from publicly available source. The model was fitted to the data using the non-linear least square method with Levenberg-Marquardt optimization.

*Results:* We analyzed 594 CTA images (4 studies with median size of 144 slices, IQR [134; 158.5]; 1:1 normal:pathology balance). Goodness-of-fit was proved by Wilcox test (p-value > 0.05 for all cases). The proposed model correctly simulated normal blood flow and hemodynamics disturbances caused by local abnormalities (aneurysm, thrombus and arterial branching).

*Conclusions:* Proposed approach can be useful for personalized CA modeling of vessels, improvement of CTA image processing and preparation of synthetic CT training data for artificial intelligence.

## 1. Introduction

Computed tomographic angiography (CTA) is the gold standard of preoperative diagnosis of abdominal aortic pathologies [1,2]. Data from routine CTA studies of the aorta are typically used to determine the vessel's size and other geometric properties [3]. Although it commonly receives far less attention, the clarity of the CT picture allows for a detailed radiographic analysis of the contrast agent distribution in the lumen.

The intravenous contrast agents (CAs) used for angiographic studies are biocompatible, chemically stable, non-ionic, isoosmolar iodine compounds [4]. According to the Beer-Lamber law [5, 6], there is a direct relationship between X-ray absorption and mixture components concentration. Based on this, and given the lack of chemical interactions between blood and CA, it is reasonable to assume that the CA component in the attenuation signal can be isolated. CA spatial distribution modeling is important for CTA planning and optimization [7, 8]. Currently, however, only perfusion procedures are used in such research [9, 10]. Another potential application of CA distribution data is the generation of synthetic non-contrast CT images [11, 12], the demand for which was highlighted in our previous study in the context of AI-opportunistic screening of aortic pathologies [13].

We propose a biorelevant model of CA spatial distribution in the aortic lumen during routine, non-perfusion CTA study. The model was validated using case studies of normal blood flow and blood flow profiles with local abnormalities (aneurysm, thrombus and arterial branching). The main model's practical applications are personalized CA distribution simulation and synthetic non-contrast CT data generation.

## 2. Methods

We divided our research into three stages (Figure 1). First, we hypothesized that the CT-density signal has a CA-induced deterministic component (CAiDC). Then, we designed a biorelevant mathematical model for CAiDC approximation at CTA. Finally, we prepared CTA patient data using 3D Slicer software [14] version 3.0.2 and validated the proposed model. Data processing and statistical analysis of obtained results was performed using R tool [15] version 4.2.1 with dplyr [16], RNIfTI [17], minipack.lm [18] and reshape [16] packages.

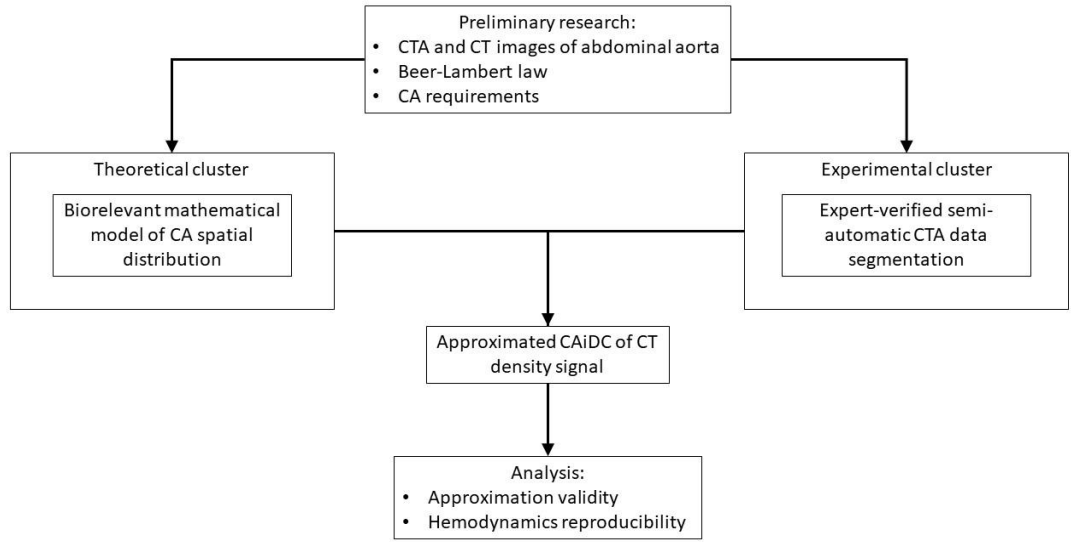

Figure 1 Study design

## 2.1 Preliminary research: CAiDC of CT-density signal

The comparison of 2D signals of the abdominal aorta at CT images of non-contrast (Figure 2, blue) and contrast (Figure 2, green) phases demonstrate the occurrence of local increase in the baseline of fluctuations, preserving the amplitude of the latter. The signal at the CTA has a rectangular shape with smoothed edges and a noisy plateau (Figure 2, shaded). The amplitude of signal depends on CA absorption, which is ranged between 100 – 500 HU [19]. Our preliminary studies have shown the reproducibility of mentioned shape in all scan directions (axial, frontal, and sagittal). Therefore, we assumed that the CA-induced component in the CT signal density is of deterministic type, and thus, it can be described with mathematical function [20].

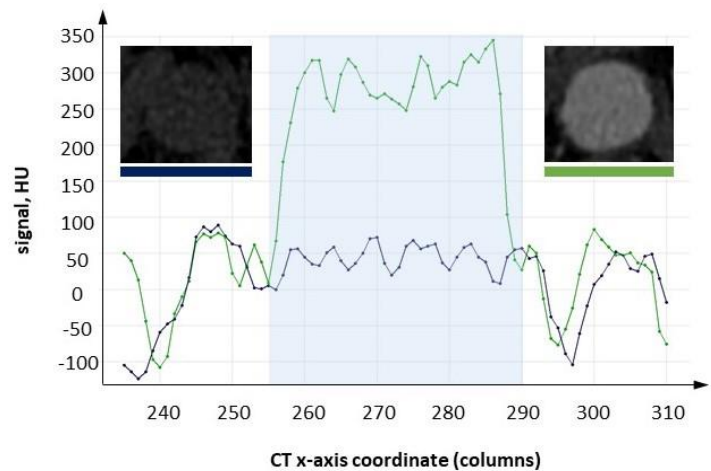

Figure 2 Signal profile at the central row of abdominal aorta (shaded) at CT (dark blue) and CTA (green) images.

## 2.2 Theoretical cluster: CAiDC mathematical model[1]

To approximate the CAiDC signal, we propose an edged plateau function (Figure 3), which is mathematically represented by the superposition of two symmetric about the x-axis and mutually offset sigmoids with a symmetric convex-concave shape:

$$F(x) = F_0 - a * (\frac{1}{1+exp(bx-c)} - \frac{1}{1+exp(dx-e)}), \qquad (2)$$

where $F_0$ is a basic signal level, $a$ is signal's amplitude, $b$ and $c$ determine rising edge slope and its inflection coordinate, $d$ and $e$ determine falling edge slope and its inflection coordinate. The coefficients $a, c, e$ are $\geq 0$; $b$ and $d$ are $> 0$; $F_0$ has no sign restriction.

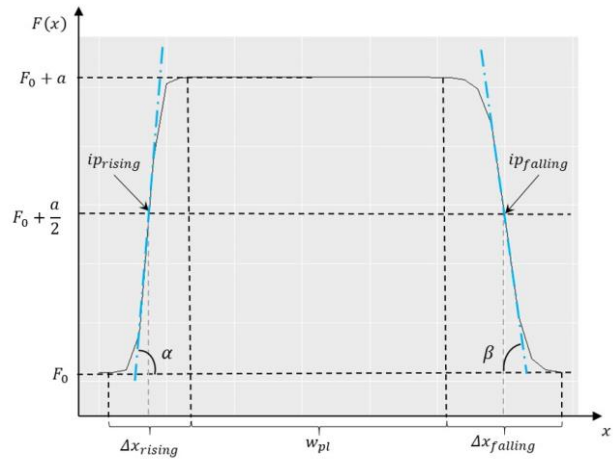

Figure 3 Model curve for CAiDC approximation. For the coefficient designations, see the text below.

It should be noted that the curves used are similar to but not identical to the Fisher-Pry symmetric sigmoid [21]. The suggested superposition makes it possible to consistently determine the exponents' arguments. Furthermore, by employing symmetry about the x-axis, it is possible to avoid correcting the amplitude of the final signal, allowing the coefficients $a$ and $F_0$ to directly define the signal's amplitude and baseline, respectively.

---

[1] Appendix A contains details on the equations and relationships used in this section.

By adjusting the exponent coefficients ($b, c, d, e$) of the two curves, we can independently control the shape of the signal's rising and falling edges. In particular, the coordinates of the inflection point for rising ($ip_{rising}$) and falling ($ip_{falling}$) edges are determined as:

$$\begin{bmatrix} ip_{rising}\left(\frac{c}{b}; F_0 + \frac{a}{2}\right) \\ ip_{falling}\left(\frac{e}{d}; F_0 + \frac{a}{2}\right) \end{bmatrix} \tag{3}$$

We used two metrics to characterize edges: the slope and the transition zone between the plateau and the baseline. To estimate edge slope, which corresponds to tangent of tilting angle of approximating line (Figure 3, blue) at the inflection points, we used the first derivative value, calculated at this point:

$$\begin{bmatrix} tg(\alpha) = \left.\frac{df}{dx}\right|_{x=ip_{rising}} = \frac{a*b}{4} \\ tg(\beta) = \left.\frac{df}{dx}\right|_{x=ip_{falling}} = -\frac{a*d}{4} \end{bmatrix} \tag{4}$$

Because of the asymptomatic behavior of sigmoids at the base and plateau levels, precise calculation of the transition zone width ($\Delta x$) is impossible. To estimate $\Delta x$, the calculation accuracy was set to 1 HU scaled by the signal's amplitude. Taking typical expected $a$ values of 100 – 500 HU [19] into account, we define calculation accuracy ($\delta y$) as:

$$\delta y = \frac{1}{\max(a)} = \frac{1}{500} = 0.002 \tag{5}$$

With this assumption, $\Delta x$ values are estimated as:

$$\begin{bmatrix} \Delta x_{rising} = 2 * \frac{\theta}{b} \approx \frac{12.42}{b} \\ \Delta x_{falling} = 2 * \frac{\theta}{d} \approx \frac{12.42}{d} \end{bmatrix}, \tag{6}$$

where $\theta$ is the constant, which depends on the calculation accuracy (see Appendix A):

$$\theta = |\ln(\delta y)| \tag{7}$$

For $\delta y$ of 0.002, the value of $\theta$ is $\approx 6.21$.

It should be noted that the aorta is surrounded by fat tissue, which can cause a sudden signal drop to negative values. In that case, $F_0$ would be $< 0$ and while the slope remain to be correct, the transition zone would include a segment of surrounding fat tissue, artificially expanding ROI (see

Figure 2 to the right of the shaded area). To address this, we used an additional condition for calculated transition zone coordinates: if the signal value at the calculated Δx endpoint was non-positive, we shifted the endpoint toward the ROI center until CAiDC became completely positive.

Exponents' coefficients also allow adjusting the width of the plateau ($w_{pl}$, Figure 3), which is equal to the difference between the inflection points' coordinates (3) with the exception of halves of the transition zones (6):

$$w_{pl} = \frac{e}{d} - \frac{c}{b} - \theta\left(\frac{1}{b} + \frac{1}{d}\right) \approx \frac{be - cd - 6.21(b+d)}{bd} \tag{8}$$

The maximum value of the plateau width across the slice, together with the transition zone of both edges, corresponds to the model-estimated vessel diameter, which is useful for vessel-muscle contact (see Figure 2 to the left of the shaded area). It is also possible to estimate the ROI by the exception of $F_0$ from the result.

The variation of model coefficients provides the curve shape flexibility (Figure 4).

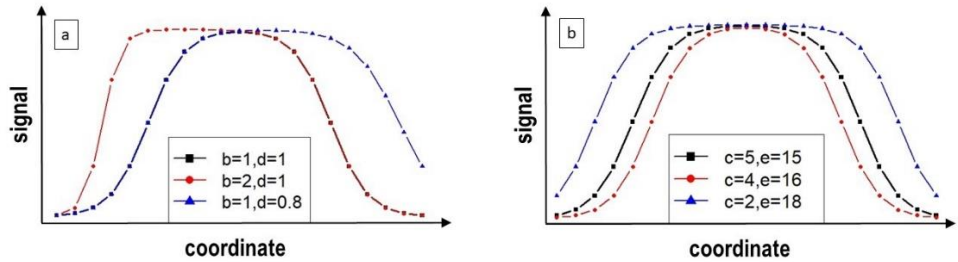

Figure 4 Model curve shape for different (a) edge slope and (b) inflection coefficient values (varied values are shown, fixed values are: $F_0 = 0$, $a = 1$, $b = 1$, $d = 1$, $c = 5$, $e = 15$)

The model operates with two-dimensional signal; however, the ROI is a 3D signal that can be described as a matrix of pixels:

$$F_{CAiDC}{}^{3D} = \left\|(F_{CAiDC})_{i,j}\right\|_{row \times column} \tag{9}$$

To avoid bias, data should be processed in both frontal (by rows, *i*) and sagittal (by columns, *j*) directions. This is significant because approximating only by one direction risks suppressing local patterns in the other one. Figure 5 depicts an example of signal approximation for an along-row-oriented object: the resulting value depends on the processing direction.

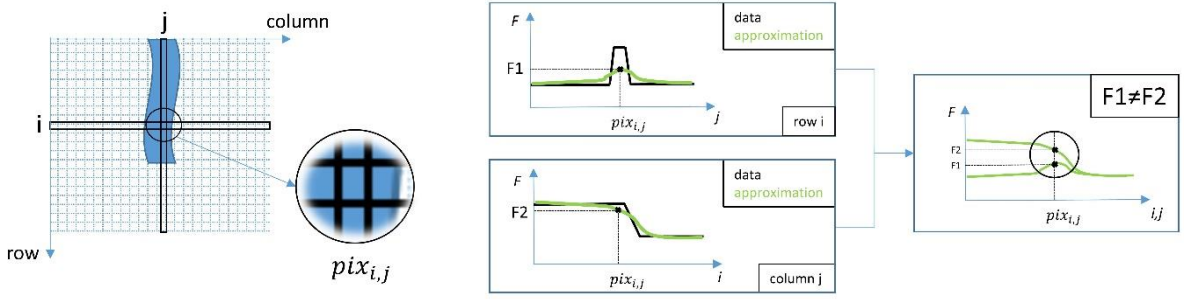

Figure 5 An example of direction-related approximation bias. Due to different contribution of the signal in the row and column directions, there is a difference in the resulting approximated amplitudes F1 and F2.

In our approach, we calculate the resulting value of ($F_{CAiDC}(pix)$) at each point by the criteria of the difference minimization between the original value $F(pix)$ and the approximated one by the corresponding row $F_i(pix)$ and column $F_j(pix)$:

$$F_{CAiDC}(pix) = \begin{cases} F_i(pix), if\ |F_i(pix) - F(pix)| \leq |F_j(pix) - F(pix)|, \\ F_j(pix), if\ |F_i(pix) - F(pix)| > |F_j(pix) - F(pix)|. \end{cases} \quad (10)$$

The suggested method (Equation 10) uses a pointwise reproduction of the three-dimensional CAiDC function to optimize the model by avoiding the input of multiple coefficients governing the local effects.

*2.3 Theoretical cluster: model biorelevance*

Each of listed in section 2.2 metrics were introduced to assess particular biophysical processes:

1. The $F_0$ coefficient represents the CT-density of the area surrounding the vessel. This baselevel value, although not being directly related to vessel processes, allows the model to specify the edge slope.

2. The $a$ coefficient corresponds to CA CT-density. Its value at each point creates a pattern that represents the flow profile at the particular slice, reflecting flow heterogeneity (if present).

3. Edge slope, alone or paired with the transition zone ($\Delta x$), defines the intensity of near-

wall and in-wall hemodynamic processes on the vessel's diametric (opposite) sides.

4. The width of plateau ($w_{pl}$) provides the estimate of the flow core area.

Each point of the final signal is located at the intersection of the approximated row and column and its value is selected based on the proximity to the real data (Equation 10). Hence, the final signal $F_{CAiDC}$ is a function defined by a double set of coefficients $f(F_{0i}, a_i, b_i, c_i, d_i, e_i, F_{0j}, a_j, b, c_j, d_j, e_j)$. For the plateau area, $F_{CAiDC}$ amplitude is selected from the pair $(F_{0i} + a_i, F_{0j} + a_j)$, depending on the proximity criteria. The edge area characteristics (slope and transition zone width) are calculated not by-point, but for the entire edge region. We assumed that the edge region approximation results are similar between rows and columns, which allows one to define coefficients $b,c,d,e$ as row or column approximated-only. By default, we chose the row values. According to this, Equation 10 can be written as:

$$F_{CAiDC} = \begin{cases} f(x, F_{0i}, a_i, b_i, c_i, d_i, e_i), & if\ x \in \text{transition zone}, \\ f(x, F_{0i}, a_i, F_{0j}, a_j), & \text{otherwise.} \end{cases} \quad (10.1)$$

To examine this assumption we performed Kruskal–Wallis test [22] of the result of by-pixel edge region approximation and those approximated by rows and columns. In cases of statistically significant difference (*p*-value < 0.05) we performed the Dunn's test [23] to define, which direction (row or column results) to choose.

To validate the proposed model's biorelevance, we examined its ability to reproduce key hemodynamic processes: general flow profile and its local effects.

*2.3.1 Flow profile*

Blood has a flattened parabolic flow profile as a non-Newtonian fluid [24] which is well-consistent with the proposed CAiDC (see Figures 3 and 5).

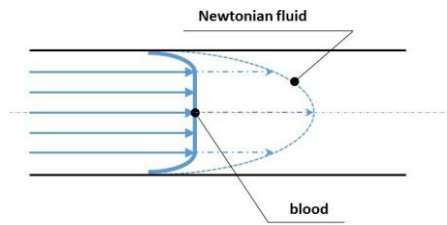

Figure 5 Blood flow profile

An obvious approach for aneurysm detection is the diameter measurement [1]. However, aneurysm changes not only the aortic diameter, but also the blood flow profile. Turbulent flows, which are common in large aneurysms [25], cause additional profile flattening [26]. This effect can be measured using the transition zone-to-plateau width ratio, which is expected to be lower for normal lumen. The edge slope can also differ between norm and aneurysm, being steeper for the former.

To reliably capture the core flow area, we localized of the vessel geometric center at each slice. For the central row and column, we extracted values of transition zone widths, plateau width, and slope angle tangent for both edges. To avoid confounding associated with flow heterogeneity, we chose non-curtosis and magistral-vessel-free aortic segments only. To obtain reproducible results, we compared calculations of at least ten slices. Slices of normal flow were extracted from the vessel with diameter < 25 mm. As aneurismal ones we extracted slices for which vessel diameter was ≥ 30 mm [1]. The Mann-Whitney-Wilcoxon test [27] (U-test) of the edge slope and $\Delta x/w_{pl}$ ratio was used to assess the model's ability to distinguish between normal and aneurismal lumen.

### 2.3.2 CA near-wall hemodynamics

We investigated near-wall flow pattern at region of vessel branching and distinct thrombosis as these areas are expected to demonstrate local heterogeneity [28, 29] To estimate model's sensitivity in detecting flow changes due to vessel branching we compared data of slices with and without magisterial vessel by Kruskall-Wailes test. We also compared within-slice transition zone width between areas with direct flow-wall contact and with thrombus-wall barrier by paired U-test. The former is expected to have a thinner near wall layer, and therefore, a steeper slope. In contrast, the transition zone should be wider in blood-to-clot contact. To do that, we chose row or column, which included both thrombosed and normal wall.

*2.4    Experimental cluster: CTA data segmentation*

We used the open source dataset [30] containing CTA studies with and without signs of aortic aneurysm in DICOM format. Inclusion criteria were as follows: arterial phase studies with the slice thickness equal to 1 mm or less. We did not exclude cases with thrombotic masses, atherosclerosis or calcinates as such findings are typical complications of aortic aneurysms [31]. In case of calcinates, the absorption amplitude of corresponding areas was downscaled to the value of 60% of maximal lumen's signal amplitude at each slice, since calcinates are etiologically immersed into the vessel wall [32]. The 60% level was chosen by our preliminary research results, which showed about 40% absorbance difference between the aorta center and walls areas.

The target region of interest (ROI) in our study was the aortic vessel, which consists of the lumen and its wall (Figure 6a). While the lumen boundaries can be precisely visually defined, the boundaries of the vessel's wall are ambiguous. Therefore, we used only aortic lumen (Figure 6b, blue) as a target for manual segmentation (S-region). To define a baselevel signal for our model, (see section 2.3) we used propagated region (P-region), which includes lumen, its wall and a thin area of surrounding tissues (Figure 6c, red).

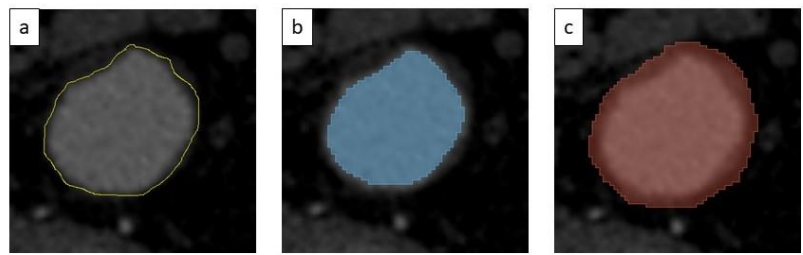

Figure 6 Example of ROI (a), S-region (b) and P-region (c) at the axial projection of the aortic vessel

To obtain S-region, a technician performed initial semi-automated area segmentation using 3D Slicer software with either "grow from seed" or "fill between slice" tool. A radiologist with three years of relevant experience verified the prepared data slice by slice, with manual mask correction in the case of disagreement with the initial segmentation. For the P-region, we performed S-region mask propagation to include aortic wall and area of reliable CA absence.

Propagated region was obtained by radial expansion of the S-region mask with a radius R:

$$R = 2 * w_{th} * w_{pix}, \quad (11)$$

where $w_{th}$ is the thickness (mm) of the abdominal aortic wall, $w_{pix}$ is the pixel density (px/mm). The value of the radius was chosen to include both wall and surrounding tissues. Vascular wall thickness was determined as 2 mm, while resolution of CT scans is usually around 1–2 pixel per mm [33]. Thus, expected value of R ranged between 4 and 8 pixels. We used NIfTI [34] as an export format for the segmentation results.

## 2.5 CAiDC approximation

To obtain CAiDC we performed real data approximation via nonlinear least square (NLS) method [35, 36] to minimize the difference between observed and model values. For automatic calculation we used the nlsLM function, where LM stands for the Levenberg-Marquardt algorithm that allows switching from the default steepest descent approach to the Gauss-Newton algorithm, which is more robust to non-optimal initial values [37].

The initial values for approximation are data-driven and have a significant impact on the results. We chose the following initial conditions at P-region of each slice:

1. $F_0$ is the minimum signal value;
2. *a* is the difference between maximal amplitude value and minimal positive signal;
3. *c* and *e* are coordinates of the start and end of each line of the P-region mask;
4. *b* and *d* are equal to 1. We chose the case of a symmetric curve as initial one, as these coefficients have indirect relationships with the signal shape (see Figure 3).

The resulting values for each coefficient are obtained iteratively by the nlsLM function. All coefficients are updated simultaneously until convergence is achieved [38].

There is no commonly used metric for estimating the validity of NLS approximation. However, there are several methods and techniques for evaluating its quality, such as statistical tests of comparing predicted values to real data or calculating residuals. We used results of U-test for real and approximated data as a major metric to assess slice by slice goodness-of-fit. We also calculated root mean square error (RMSE) as a descriptive metric.

## 3. Results

### 3.1 Data characteristics

The total number of processed CT images was 594 (Table 1).

| Parameter | Value |
| --- | --- |
| Name of procedure | abdominal computed tomography |
| Target pathology | aortic aneurysm (ICD 10 code: I71) |
| Scan gathering period: beginning/end | 08.01.2022/31.01.2022 |
| Number of CT scans/patients | 4/4 |
| Female, % | 50 |
| Age (years): min/median/max | 40/47/55 |
| Data parameters | |
| Type of marking | binary |
| Class assignment criteria | Pathology was considered when the largest diameter of the abdominal aorta in the axial plane was: 1) 25 – 29 mm for dilatation; 2) ≥ 30 mm for aneurysm. Cases with largest diameter of the abdominal aorta in the axial plane < 25 mm were treated as norm. |
| Number of cases containing arterial/native phase scans | 4/4 |
| Slice thickness (mm) | ≥ 1 |
| Number of slices per study (1; 2; 3; 4) | 178; 152; 128; 136 |
| Study class; maximal aortic diameter (mm) | aneurysm; 34 mm<br>dilatation with thrombosis; 27 mm<br>norm with multiple calcification; 20 mm<br>norm; 16 mm |

Table 1 Dataset composition

The calculated radius for P-region (Equation 9) of 4 pixels reliably captured surrounding non-vessel area for all studies (Figure 7).

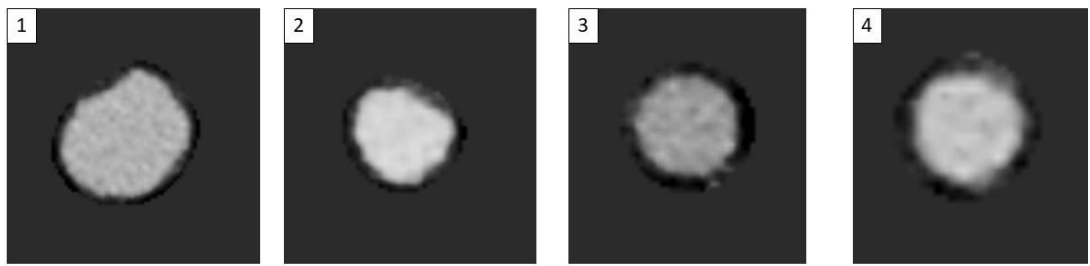

Figure 7 Examples of P-region data (the picture number corresponds to the case number)

*3.2 CAiDC approximation validity*

The U-test results of statistical comparison between approximated CAiDC and real data at each slice are presented in the Figure 8. For all studies there were no statistically significant differences between the approximated and real data.

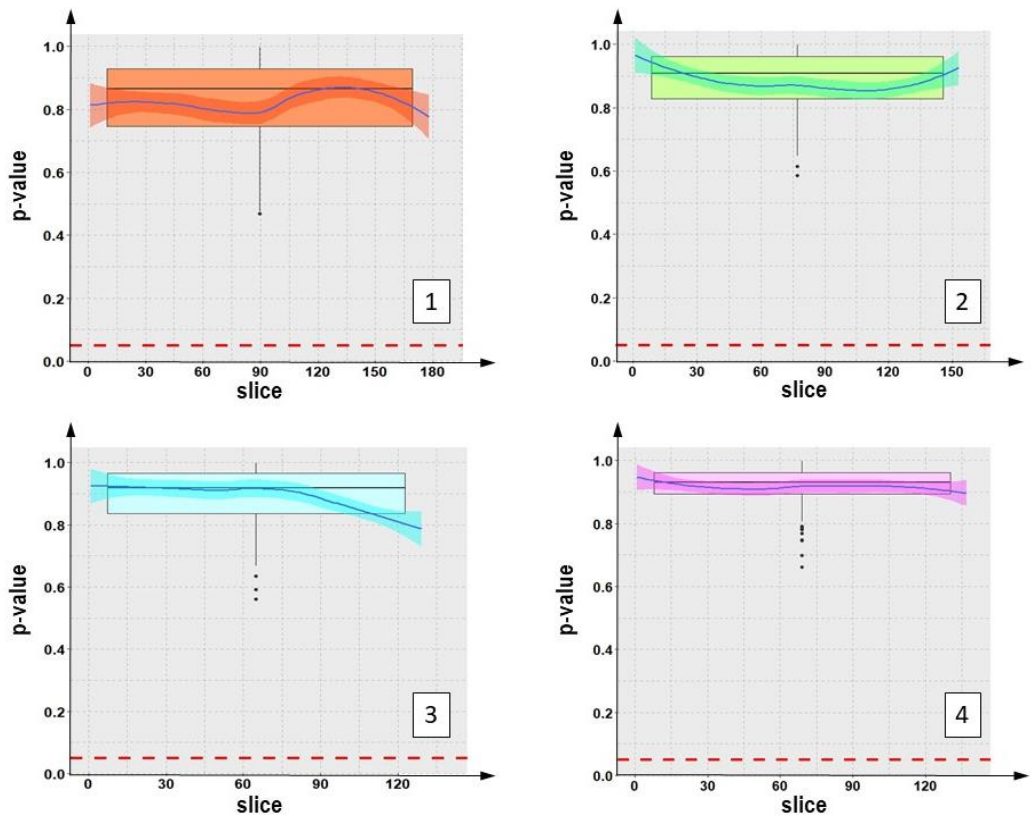

Figure 8 Approximation quality of the CAiDC model. The panel number corresponds to the case number. Boxplots represent U-test results (*p*-value) for approximated CAiDC and real data. Curve shows smoothed *p*-value distribution with 95% CI; red dashed line denotes *p*-value 0.05.

The RMSE results of the per-slice comparison of approximated and real data (Figure 9) ranged from 10 to 20 HU with the minimal values for the second case and the maximal values for the forth one. We were unable to find literature sources presenting the reference values of this metric, hence, these data may be the primary estimate of such models' goodness-of-fit.

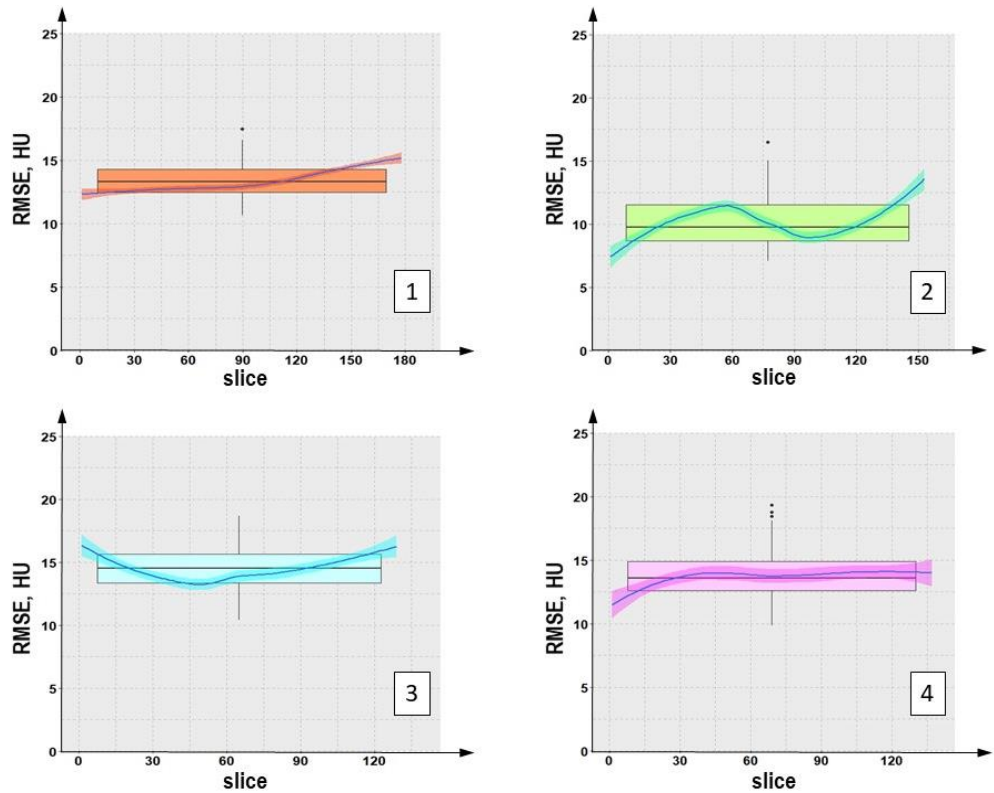

Figure 9 Goodness-of-fit estimation of the CAiDC model. The panel number corresponds to the case number. Boxplots represent RMSE values for approximated CAiDC and real data. Curve shows smoothed RMSE value distribution with 95% CI.

### 3.3   Model's biorelevance validity

First, we verified our presumption (section 2.3) on the similarity of signal approximation for the transition zone obtained by row $F_i(pix)$, by column $F_j(pix)$, and the by-point $F_{CAiDC}(pix)$. Except for the one slice of the third study (0.2 % of total set), the results of Kruskall-Wailles test supported our assumption of no statistically significant differences (Figure 10).

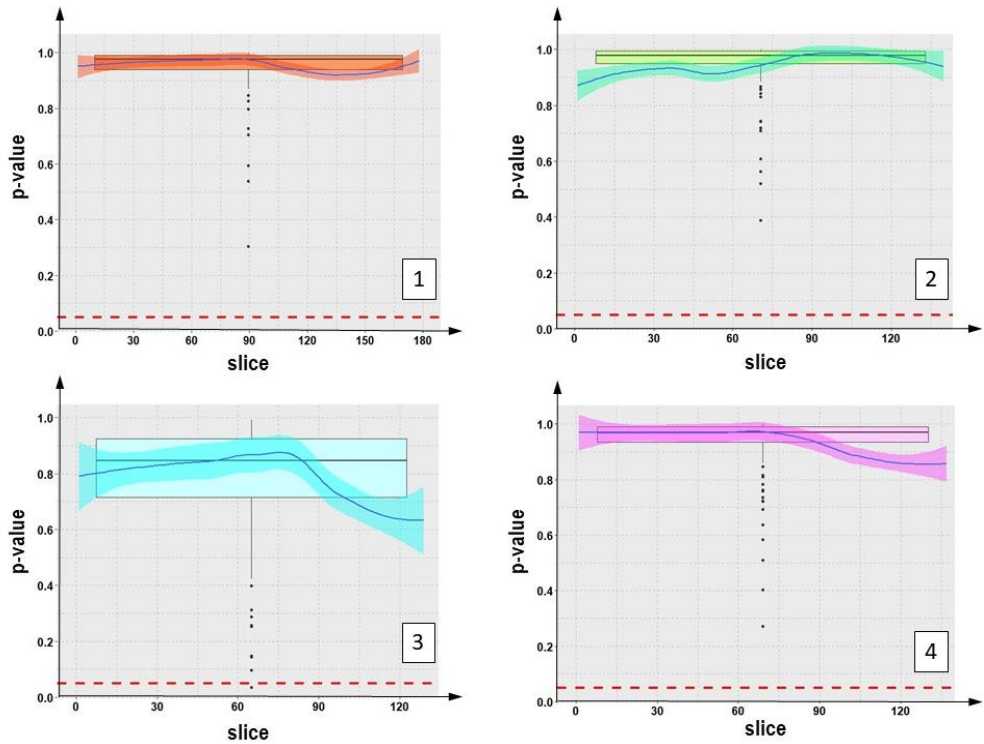

Figure 10 Edge region approximation similarity between the three approaches. The panel number corresponds to the case number. Boxplots represent Kruskall-Wailes test results (*p*-value). Curve shows smoothed *p*-value distribution with 95% CI; red dashed line denotes *p*-value 0.05.

In contrast with the well-approximated transition zone (Figure 11a), the case with poor similarity contained vessel lumen that had magisterial artery connection (Figure 11b). However, Dunn's post-hoc test revealed statistical similarity at this slice between by-column $F_j(pix)$ and by-point $F_{CAiDC}(pix)$.

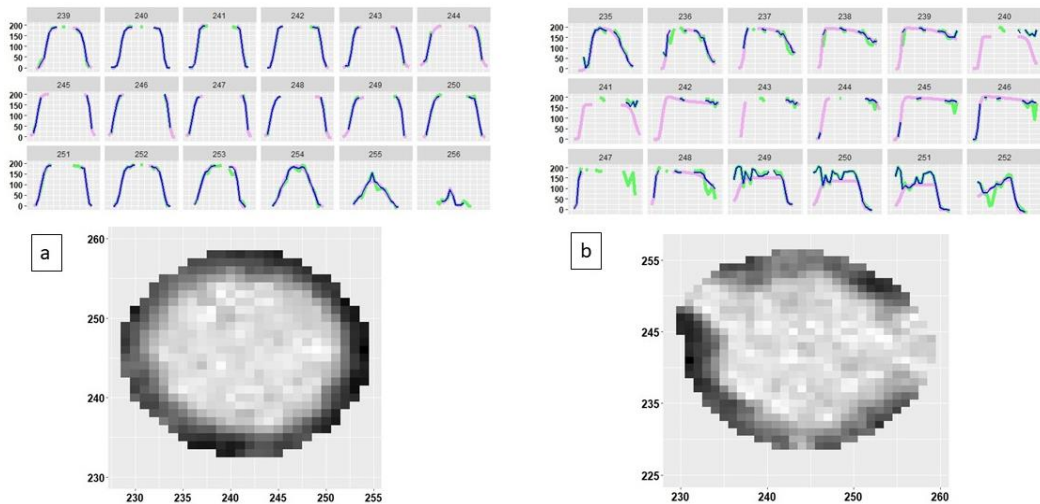

Figure 11 An example of transition zone with statistically similar (a) and non-similar (b) results

obtained by-row (green) and by-column (violet) approximation. Blue denotes by-point $F_{CAiDC}$.
Numbers on the facets correspond to the column number.

After confirming the assumption of unidirectional data approximation in the transition zone, we calculated and compared the metrics characterizing the key hemodynamic processes.

*3.3.1 Flow profile*

Cases 1 and 3 were chosen to obtain continuous non-curtosis and magistral-vessel-free aortic segments of 21 slices length (Figure 12). We extracted a row and a column at the geometric center at each of the slices and calculated the paired edge slopes and the transition zone-to-plateau width ratio.

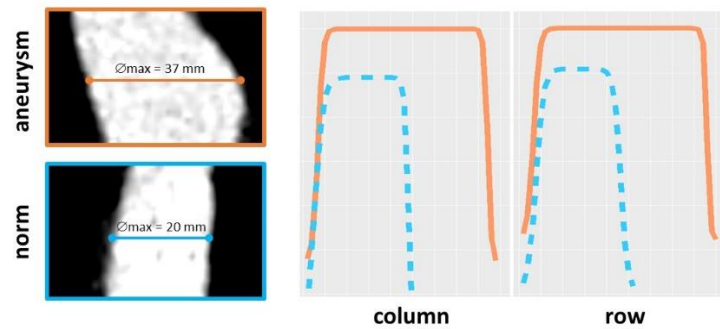

Figure 12 Extracted segments (sagittal view) and example of left-side-aligned signals from the central lines of the ROI with aneurysm (case 1, orange) and norm (case 3, light blue)

Although visual comparison of edges (Figure 12, right) failed to demonstrate any essential difference, the results of U-test between norm and aneurysm cases for transition zone-to-plateau width ratio $\Delta x/w_{pl}$ (Figure 13, 1a and 1b) and absolute slope angles (Figure 13, 2a and 2b) demonstrated statistically significant difference for both metrics (*p*-value <0.05). However, the slope angle is not a reliable metric: its values for the aneurysm and normal case show inconsistent behavior for the rising and falling edges (Figure 13 2a and 2b). As a result, when the values for both edges are combined, the U-test demonstrated no statistically significant difference (*p*-value = 0.9). Thus, $\Delta x/w_{pl}$ is more reliable metric, as accounting the behavior of the whole signal. This ratio is smaller for a flattened profile, typical to the aneurysmal lumen. The results of the U-test for this metric separately for the rising and falling edges showed *p*-value <0.0001, which retained for joint data also.

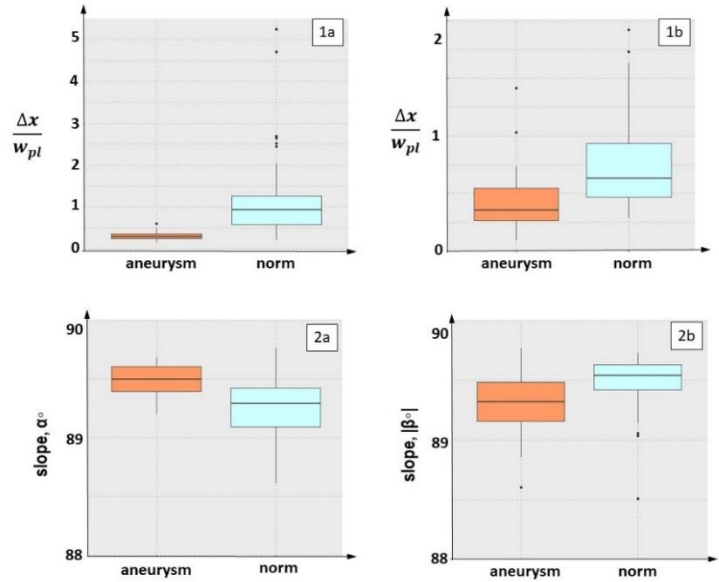

Figure 13 Aneurysmal (orange) versus normal (light blue) edge comparison. Boxplots represent transition zone-to-plateau width ratio (1) and absolute slope values (2) for the rising (a) and falling (b) edges.

Thus, proposed model is able to detect aneurysms not only by estimative diameter measurement (CAiDC signal length), but also by the flow profile changes.

*3.3.2 Detection of flow abnormalities*

Our approach of 3D signal formation (Equation 10) allowed us to accurately capture local flow patterns at the plateau region, which may be caused, for example, by vessel branching. We investigated an area of 10 slices containing magisterial vessels branching (Figure 14a). We compared a lumen-only flow (S-region) at the set, containing slices of the pre-branching, branching, and post-branching area. Kruskall-Wailes test's results (Figure 14b) showed statistically significant difference between reliably non-branched areas (Figure 12a, slices 96 and 105) and a branching one (Figure 14a, slice 99). With slices 96 and 105 excluded, data contained no statistically significant difference.

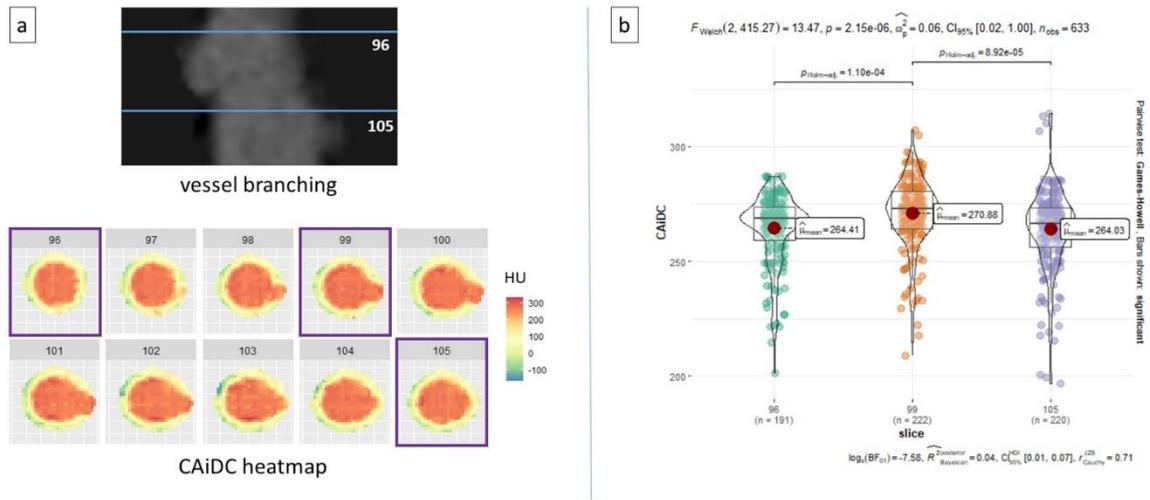

Figure 14 Example of CAiDC flow investigation. Branching segment (case 4, slices 96 – 105) heatmaps (a) and the result of within-subset comparison (b) for the highlighted slices of normal (96, 105) and branched (99) lumen are shown.

Therefore, model can be used to capture estimative axial zone of flow heterogeneity. However, to avoid misinterpretations the comparison with objective flow dynamics (obtained, for example, by MRI) is required, which was impossible in this study due to the lack of necessary data.

To investigate near-wall hemodynamic we extracted a slice from the second case with one-side-oriented thrombotic clot (Figure 15a). Clot area included 10 rows. We performed a pairwise comparison of the rising (blood-to-clot contact) and falling (blood-to-wall contact) edges of extracted rows. The results of paired U-test for the transition zone widths (Figure 15b) and slope angles (Figure 15c) showed statistically significant difference ($p$-value < 0.05). Proposed model demonstrated smoother (wider transition zone with smaller slope angle) contact than for the normal blood-to-wall contact. The possible reason of this effect is due to the possibility of clot-blood interaction [39], hence, CA-to-clot diffusion during CTA procedure.

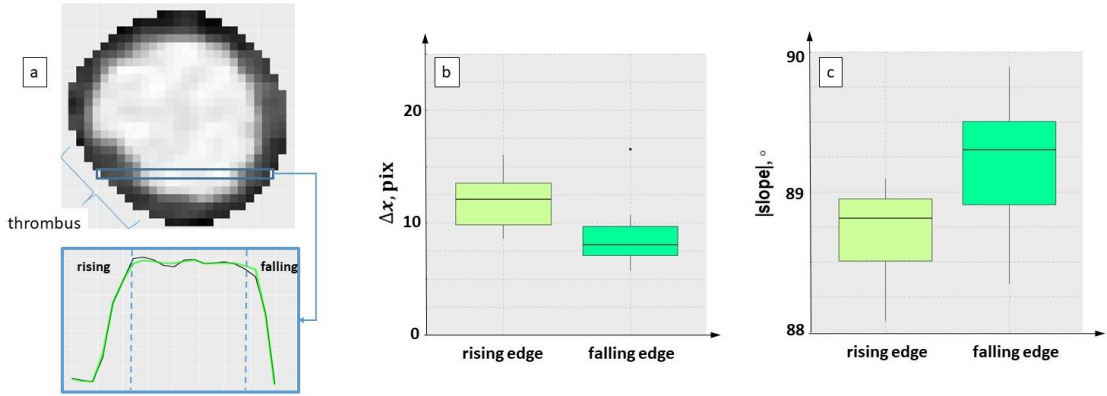

Figure 15 Difference in local hemodynamics for blood-to-thrombus (rising edge) and blood-to-wall (falling edge) contact. Boxplot of transition zone widths (b) and absolute slope angles (c) are shown.

Thus, model demonstrated ability to capture near-wall-hemodynamics difference for cases of CA-clot and CA-to-wall interaction.

### 3.4 CA elimination

We also performed CAiDC elimination, obtaining synthetic non-contrast image of the ROI (Figure 16a). This can be usable for obtaining non-contrast images with preserved mark up that was performed at contrast-phase image (figure 16b), as the vascular area is challenging to distinguish from the surrounding tissue at native phase (see Figure 2, blue curve). Although CAiDC elimination shows promising results, the out-ROI area of a CT-image should also be taken into account. There is difference in signal noise between native and arterial phase images due to the photon starvation effect [40]. Obtaining of images appropriate for AI applications requires careful calibration, which is the object of our future research.

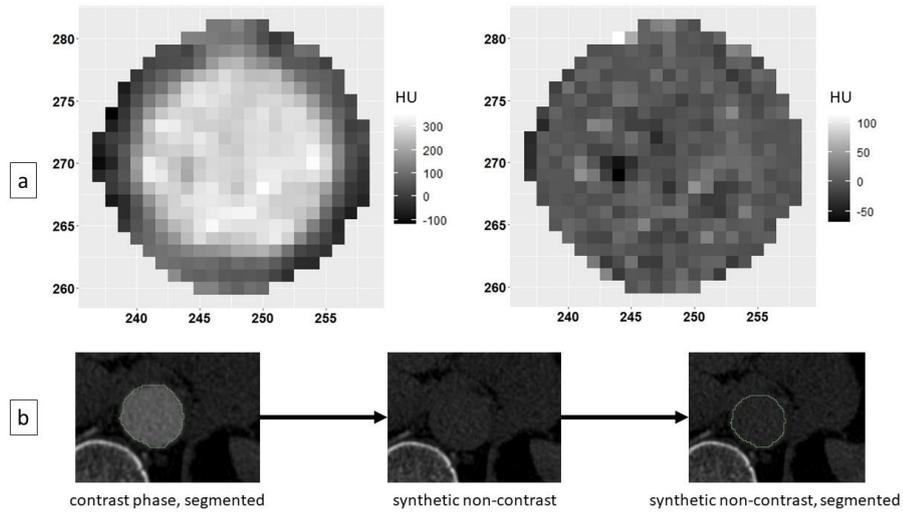

Figure 16 Example of CAiDC elimination (a) and possible pipeline of non-contrast marked CT creation (b)

## 4. Discussion

Our goal was to analyze the CA-induced deterministic component of the CT-density signal in the abdominal aortic region. To describe the deterministic component, we developed a biorelevant mathematical model. To validate the model, we prepared and processed publicly available CTA data of 4 patients, which amounted to a total of 594 images with a median set size of 144 slices (IQR [134; 158.5]; 1:1 normal:pathology balance). We used expert-verified manual mark up and automatic post-processing (Subsection 2.4) for the data segmentation. The final data set included areas of thrombosis (case 2), aneurysms (cases 1 and 2), and normal flow (cases 3 and 4).

The proposed model describes the behavior of the two-dimensional signal as an original superposition of two symmetric sigmoids (Equation 2). This allowed the flow behavior in the lumen and vessel wall to be independently simulated at the entire ROI area in the axial projection. We introduced an approach to account for the CT image's properties by correction factor (see equation 7). Using the proposed bidirectional data processing and pointwise formation of a three-dimensional signal (see Equation 10) with separate approach for edge and plateau zones (see Equation 10.1), we avoided adding redundant coefficients to the model. A cross-sectional comparison of the real and obtained data using a nonparametric Wilcoxon test showed that there was no statistically significant difference (p > 0.05) in all cases, demonstrating the accuracy of the approximation (see Figure 8). We have used an additional RMSE metric to access the goodness-of-fit of the model and proposed the RMSE threshold of 20 HU (see Figure 9) for the good agreement between the model and real data. The model includes six independent coefficients, which combination allowed us to describe and estimate both the flow core and the near-wall hemodynamics.

Biorelevance of the model was validated on the cases of aneurysm, thrombus and vessel branching. To illustrate the model's sensitivity in detecting flow core differences between normal and aneurysmal lumen (see Figure 12), we employed plateau-to-transition zone widths ratio as a reliable metric. The expected aneurysm-associated flattering of the flow core was supported by the results of U-test ($p$-value < 0.0001). We investigated local heterogeneity caused by the vessel branching by comparing CAiDC at slices of uninterrupted and branched lumen, estimating model's validity in detecting flow abnormalities areas ($p$-value < 0.05, see Figure 14). Difference in local hemodynamics was also demonstrated by the comparison of the rising and falling edge zones at the slice with area of distinct thrombus, being associated with different blood-to-wall contact (paired Wilcox test $p$-value < 0.05, see Figure 15). As a proof-of-concept, we performed elimination of approximated CAiDC to obtain synthetic non-contrast CTA images of the ROI. Proposed approach allowed us to remove CA, preserving original signal fluctuations (see Figure 16).

### 4.1 *Implications of the results for practice, policy, and future research*

Our model was primarily developed to analyze spatial CA distribution in routine CTA procedure. We assumed that this data can be used in personalized CTA procedure simulation to optimize the procedure in case of follow-up studies necessity. On the other hand, while gathering the data from multiple patients, the model can be useful to create an expected CA distribution profile. Deviations from this profile could signal of technical or methodological errors, or pathology existence. As the model provides objective numerical characterization of the hemodynamics disturbances, which cannot always be detected visually, it can be useful for medical decision-making.

Additional model's application is the CA elimination functionality. It can be useful for synthetic non-contrast phase simulation in the aortic area. We find it essential as it provides non-contrast training data with accurate segmentation obtained from "gold standard" CTA studies markup. We expect to increase the amount and quality of training data in this area to improve the development of AI-algorithms for aortic pathologies detection, including the opportunistic screening strategy.

Future research implies an in-depth study of the model applicability for personalized CTA simulation using the anthropomorphic aortic phantom that is currently being designed. Phantom study, as we expect, will allow us to simulate different left ventricular ejection fraction, and investigate different CAs. We also plan to validate the model's ability to detect vessel wall and near-wall abnormalities on a larger sample, and by comparison of model results with objective flow metrics obtained by MRI [41, 42].

Intriguing direction of the future research is diagnostics of aortic pathologies by the wall nourishment estimation. The muscular layer of the aorta under physiological conditions is nourished by two sources: direct diffusion from luminal blood flow [43] and by *vasa vasorum* [44]. As a result, CA is expected to enter the vascular wall via the bloodstream. The *vasa vasorum* density is individual, but it is also reported to be the most dense in the arch aorta, decreasing distally, with the lowest density in the infrarenal abdominal aorta [45, 46]. The luminar blood diffusion, oppositely, is known to be predominantly in the infrarenal segment of the abdominal aorta [47]. Therefore, there are prospects of individual contribution estimation of these effects by comparison of the relation between widths of plateau and transition zone at slices of the suprarenal and infarenal aorta. This may be promising in detecting, for example, ischemia-specific patterns.

Finally, we plan to reconstruct obtained data in axial direction to investigate flow processes connected with previously demonstrated pulse wave propagation [48].

*4.2 Limitations*

Although our research showed promising results, the study has several limitations.

Firstly, we used artificial suppression of bright objects outside the lumen area. However, calcinates do not participate in flow process, since they are etiologically immersed into the vessel wall [22].

Secondly, our model does not directly include pulse wave component, which is expected, as average abdominal CT procedure (arterial phase) takes up to 20 seconds [2]. However, since we are processing slices in axial projection, the contribution of the pulse wave is indirectly taken into account as one of the components shaping the behavior of the flow core. Moreover, we can extract separately the flow core and near-wall layer to investigate this process, as stated above.

Finally, our data set included cases with aortic abnormalities that were not entirely representative and were not cardiac cycle synchronized. We are currently compiling a larger data set in order to examine the reproducibility of the results obtained and to broaden potential simulated effects.

## 5. Summary and conclusions

An approach of extraction and determination of CA-induced deterministic signal at aortic region at CTA is presented. We propose a biorelevant mathematical model of spatial CT-density distribution at the aortic lumen and wall. The model performance has been validated using real-patient CT data. Model is able to accurately distinguish flow abnormalities connected with arterial branching, and to reveal flow-specific signs of aneurysm and thrombosis.

## CRediT authorship contribution statement

approval of the version to be submitted. **Yuriy A. Vasilev, Anton V. Vladzimirskyy** and **Olga V. Omelyanskaya:** Project administration, Resources, Final approval of the version to be submitted.

## Declaration of competing interest

The authors declare that they have no known competing financial interests or personal relationships that could have appeared to influence the work reported in this paper.

## Data availability

Data containing S-region and P-region segmentation binary label maps are available upon request.

## Acknowledgements

The authors would like to thank Alexander V. Soloviev for valuable assistance in expert data segmentation.

## Funding

This research was funded in accordance with the Order No. 1196 dated December 21, 2022 "On approval of state assignments funded by means of allocations from the budget of the city of Moscow to the state budgetary (autonomous) institutions subordinate to the Moscow Health Care Department, for 2023 and the planned period of 2024 and 2025" issued by the Moscow Health Care Department (EGISU) [grant number 123031500002-1].

## References

[1] Erbel R, Aboyans V, Boileau C, Bossone E, Di Bartolomeo R, Eggebrecht H, et al. 2014 ESC Guidelines on the diagnosis and treatment of aortic diseases: Document covering acute and chronic aortic diseases of the thoracic and abdominal aorta of the adult. The Task Force for the Diagnosis and Treatment of Aortic Diseases of the European Society of Cardiology (ESC). Eur Heart J 2014;35:2873–926. https://doi.org/10.1093/EURHEARTJ/EHU281.

[2] Baliyan V, Shaqdan K, Hedgire S, Ghoshhajra B. Vascular computed tomography angiography technique and indications. Cardiovasc Diagn Ther 2019;9. https://doi.org/10.21037/CDT.2019.07.04.

[3] Yu T, Zhu X, Tang L, Wang D, Saad N. Review of CT angiography of aorta. Radiol Clin North Am 2007;45:461–83. https://doi.org/10.1016/J.RCL.2007.04.010.

[4] Contrast media | Radiology Key n.d. https://radiologykey.com/contrast-media-2/ (accessed May 16, 2023).

[5] I.H. Lambert Photometria, sive, De mensura et gradibus luminis, colorum et umbrae

[microform] | National Library of Australia n.d. https://catalogue.nla.gov.au/Record/857394 (accessed May 16, 2023).

[6] Beer. Bestimmung der Absorption des rothen Lichts in farbigen Flüssigkeiten. Ann Phys 1852;162:78–88. https://doi.org/10.1002/ANDP.18521620505.

[7] Saade C, Deeb IA, Mohamad M, Al-Mohiy H, El-Merhi F. Contrast medium administration and image acquisition parameters in renal CT angiography: What radiologists need to know. Diagnostic and Interventional Radiology 2016;22:116–24. https://doi.org/10.5152/DIR.2015.15219.

[8] Hinzpeter R, Eberhard M, Gutjahr R, Reeve K, Pfammatter T, Lachat M, et al. CT Angiography of the Aorta: Contrast Timing by Using a Fixed versus a Patient-specific Trigger Delay. Radiology 2019;291:531–8. https://doi.org/10.1148/RADIOL.2019182223.

[9] Valdiviezo C, Ambrose M, Mehra V, Lardo AC, Lima JAC, George RT. Quantitative and qualitative analysis and interpretation of CT perfusion imaging. J Nucl Cardiol 2010;17:1091–100. https://doi.org/10.1007/S12350-010-9291-6.

[10] American College of Radiology, 2017. ACR–NASCI–SIR–SPR practice parameter for the performance and interpretation of body computed tomography angiography (CTA)

[11] Sandfort V, Yan K, Pickhardt PJ, Summers RM. Data augmentation using generative adversarial networks (CycleGAN) to improve generalizability in CT segmentation tasks. Sci Rep 2019;9. https://doi.org/10.1038/S41598-019-52737-X.

[12] Chen Y, Yang XH, Wei Z, Heidari AA, Zheng N, Li Z, et al. Generative Adversarial Networks in Medical Image augmentation: A review. Comput Biol Med 2022;144. https://doi.org/10.1016/J.COMPBIOMED.2022.105382.

[13] Kodenko MR, Vasilev YA, Vladzymyrskyy A V., Omelyanskaya O V., Leonov D V., Blokhin IA, et al. Diagnostic Accuracy of AI for Opportunistic Screening of Abdominal Aortic Aneurysm in CT: A Systematic Review and Narrative Synthesis. Diagnostics 2022, Vol 12, Page 3197 2022;12:3197. https://doi.org/10.3390/DIAGNOSTICS12123197.

[14] 3D Slicer image computing platform | 3D Slicer n.d. https://www.slicer.org/ (accessed May 16, 2023).

[15] Posit | The Open-Source Data Science Company n.d. https://posit.co/ (accessed May 16, 2023).

[16] Wickham H. Reshaping Data with the reshape Package. J Stat Softw 2007;21:1–20. https://doi.org/10.18637/JSS.V021.I12.

[17] Package "RNifti" Title Fast R and C++ Access to NIfTI Images 2023.

[18] Elzhov V, Mullen KM, Spiess A-N, Maintainer BB. Title R Interface to the Levenberg-Marquardt Nonlinear Least-Squares Algorithm Found in MINPACK, Plus Support for Bounds n.d.

[19] Bell D, Greenway K. Hounsfield unit. RadiopaediaOrg 2015. https://doi.org/10.53347/RID-38181.

[20] Chen W-K. The electrical engineering handbook 2005:1208.

[21] Fisher JC, Pry RH. A Simple Substitution Model of Technological Change*. TECHNOLOGICAL FORECASTING AND SOCIAL 1971;3:75–88.

[22] kruskal.test function - RDocumentation n.d. https://www.rdocumentation.org/packages/stats/versions/3.6.2/topics/kruskal.test (accessed May 16, 2023).

[23] dunn.test function - RDocumentation n.d. https://www.rdocumentation.org/packages/dunn.test/versions/1.3.5/topics/dunn.test (accessed May 16, 2023).

[24] Ku DN, Woodruff GW. BLOOD FLOW IN ARTERIES. Annu Rev Fluid Mech 1997;29:399–434.

[25] Kemmerling EMC, Peattie RA. Abdominal aortic aneurysm pathomechanics: Current understanding and future directions. Adv Exp Med Biol 2018;1097:157–79. https://doi.org/10.1007/978-3-319-96445-4_8/COVER.

[26] Yoganathan AP, Cape EG, Sung HW, Williams FP, Jimoh A. Review of hydrodynamic principles for the cardiologist: applications to the study of blood flow and jets by imaging techniques. J Am Coll Cardiol 1988;12:1344–53. https://doi.org/10.1016/0735-1097(88)92620-4.

[27] wilcox.test function - RDocumentation n.d. https://www.rdocumentation.org/packages/stats/versions/3.6.2/topics/wilcox.test (accessed May 16, 2023).

[28] Williams AR, Koo BK, Gundert TJ, Fitzgerald PJ, LaDisa JF. Local hemodynamic changes caused by main branch stent implantation and subsequent virtual side branch balloon angioplasty in a representative coronary bifurcation. J Appl Physiol 2010;109:532–40. https://doi.org/10.1152/JAPPLPHYSIOL.00086.2010/SUPPL_FILE/DATASUPP.PDF.

[29] Chiu JJ, Chien S. Effects of disturbed flow on vascular endothelium: Pathophysiological basis and clinical perspectives. Physiol Rev 2011;91:327–87. https://doi.org/10.1152/PHYSREV.00047.2009/ASSET/IMAGES/LARGE/Z9J0011125710023.JPEG.

[30] MosMedData КТ с признаками аневризмы аорты тип III - наборы данных в лучевой диагностике n.d. https://mosmed.ai/datasets/mosmeddata-kt-s-priznakami-anevrizmi-aorti-tip-iii/ (accessed May 16, 2023).

[31] Aggarwal, S., Qamar, A., Sharma, V., and Sharma, A., 2011. Abdominal aortic aneurysm: A comprehensive review. Experimental and Clinical Cardiology, 16(1), 11 https://pubmed.ncbi.nlm.nih.gov/21523201/ (accessed May 16, 2023).

[32] Sage AP, Tintut Y, Demer LL. Regulatory mechanisms in vascular calcification. Nature Reviews Cardiology 2010 7:9 2010;7:528–36. https://doi.org/10.1038/nrcardio.2010.115.

[33] Haller SJ, Azarbal AF, Rugonyi S. Predictors of Abdominal Aortic Aneurysm Risks. Bioengineering 2020, Vol 7, Page 79 2020;7:79. https://doi.org/10.3390/BIOENGINEERING7030079.

[34] Knipe H, Moore C. NIfTI (file format). RadiopaediaOrg 2019. https://doi.org/10.53347/RID-72562.

[35] Kelley C. T., 1999 Iterative methods for optimization. – Society for Industrial and Applied Mathematics

[36] Emmert-Streib F, Moutari S, Dehmer M. Mathematical Foundations of Data Science Using R. Mathematical Foundations of Data Science Using R 2020:1–414. https://doi.org/10.1515/9783110564990/EPUB.

[37] Mohammad H, Waziri MY, Santos SA, Mohammad H, Waziri MY, Santos SA. A brief survey of methods for solving nonlinear least-squares problems. Numerical Algebra, Control and Optimization 2019;9:1–13. https://doi.org/10.3934/NACO.2019001.

[38] Remko Duursma n.d. https://www.remkoduursma.com/post/2017-04-24-nlshelper/ (accessed May 16, 2023).

[39] Munn LL, Dupin MM. Blood cell interactions and segregation in flow. Ann Biomed Eng 2008;36:534–44. https://doi.org/10.1007/S10439-007-9429-0/FIGURES/8.

[40] Mori I, Machida Y, Osanai M, Iinuma K. Photon starvation artifacts of X-ray CT: Their true cause and a solution. Radiol Phys Technol 2013;6:130–41. https://doi.org/10.1007/S12194-012-0179-9/FIGURES/9.

[41] Sughimoto K, Shimamura Y, Tezuka C, Tsubota K, Liu H, Okumura K, et al. Effects of arterial blood flow on walls of the abdominal aorta: distributions of wall shear stress and oscillatory shear index determined by phase-contrast magnetic resonance imaging. Heart Vessels 2016;31:1168–75. https://doi.org/10.1007/S00380-015-0758-X/FIGURES/7.

[42] Kobelev E., Pak N.T., Bobrikova E.E., Ussov W.Y., Kliver E.E., Sirota D.A., Chernyavskiy A.M., Bergen T.A. Diagnostic challenge: innovative approach in use of magnetic resonance imaging in aortic aneurysm // Digital Diagnostics. - 2022. - Vol. 3. - N. 3. - P. 332-339. https://doi.org/10.17816/DD108404

[43] Tanaka H, Unno N, Suzuki Y, Sano H, Yata T, Urano T. Hypoperfusion of the Aortic Wall

Secondary to Degeneration of Adventitial Vasa Vasorum Causes Abdominal Aortic Aneurysms. Curr Drug Targets 2018;19:1327–32. https://doi.org/10.2174/1389450119666180122154409.

[44] Heistad DD, Marcus ML. Role of vasa vasorum in nourishment of the aorta. Blood Vessels 1979;16:225–38. https://doi.org/10.1159/000158209.

[45] Sano M, Unno N, Sasaki T, Baba S, Sugisawa R, Tanaka H, et al. Topologic distributions of vasa vasorum and lymphatic vasa vasorum in the aortic adventitia – Implications for the prevalence of aortic diseases. Atherosclerosis 2016;247:127–34. https://doi.org/10.1016/J.ATHEROSCLEROSIS.2016.02.007.

[46] White HJ, Borger J. Anatomy, Abdomen and Pelvis, Aorta. StatPearls 2019.

[47] Grossmannova K, Barathova M, Belvoncikova P, Lauko V, Csaderova L, Tomka J, et al. Hypoxia Marker Carbonic Anhydrase IX Is Present in Abdominal Aortic Aneurysm Tissue and Plasma. International Journal of Molecular Sciences 2022, Vol 23, Page 879 2022;23:879. https://doi.org/10.3390/IJMS23020879.

[48] Mynard JP, Smolich JJ. One-Dimensional Haemodynamic Modeling and Wave Dynamics in the Entire Adult Circulation. Annals of Biomedical Engineering 2015 43:6 2015;43:1443–60. https://doi.org/10.1007/S10439-015-1313-8.

# Appendix A. Equations

*A.1 Model curve: x-symmetry versus y-symmetry*

This section provides step-by-step derivation of the proposed model's equations. As been stated, we propose an original combination of Fisher-Pry-like sigmoids. If we take two Fisher-Pry sigmoids and make then countercurrent by changing sign of exponential argument preserving their positive-defined values (Figure A1a, light blue and green), we will obtain a superposition with plateau level equal to double amplitude (Figure A1a, orange). Therefore, we will need to subtract $a$ value from the result (Figure A1a, black dashed):

$$f(x) = F_0 - a + \frac{a}{1+\exp(-bx-c)} + \frac{a}{1+exp(dx-e)} \qquad (A1)$$

However, if instead of changing exponential argument sign, we use x-axis symmetry by changing the sign of a whole fraction, we will obtain amplitude compensation (Figure A1b, orange), leading to desired curve shape. So, that we can directly use $F_0$ and $a$ to define baselevel and plateau amplitude respectively. Moreover, all the coefficients will be positive, which simplifies not only the calculation procedure, but also their biophysical interpretation. Thus, we used two sigmoids:

1) y-negative, which defines the rising edge (Figure A1b, light blue):

$$f(x) = F_0 - \frac{a}{1+exp(bx-c)} \qquad (A2)$$

2) y-positive, which defines the falling edge (Figure A1b, green):

$$f(x) = F_0 + \frac{a}{1+exp(dx-e)} \qquad (A3)$$

to obtain the following function (Figure A1b, orange):

$$f(x) = F_0 - \frac{a}{1+exp(bx-c)} + \frac{a}{1+exp(dx-e)} \qquad (A4)$$

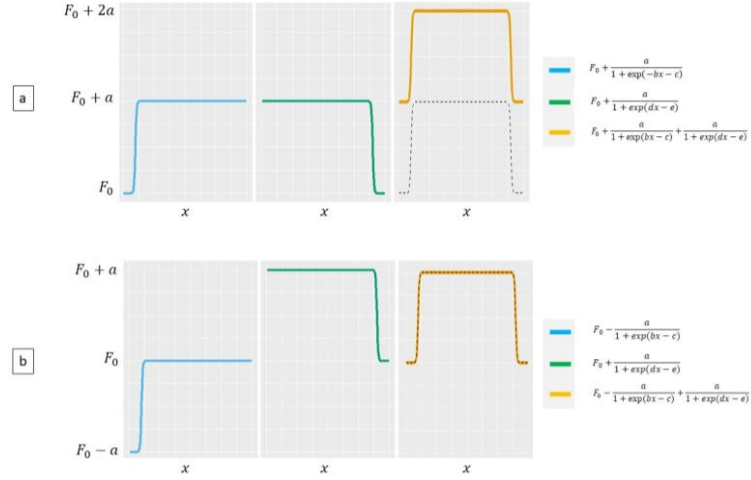

Figure A1. Model curve formation: direct superposition of primary Fisher-Pry sigmoids (a), proposed superposition (b); light blue is for rising edge sigmoid, green is for falling edge sigmoid, orange is for the superposition, black dashed line is for desired result

*A.2 Transition zone characteristics: inflection point*

The sigmoid function can be written in general form as:

$$f(x) = \frac{1}{1+exp(\varepsilon x + \vartheta)} \qquad (A5)$$

The exponential argument determines the shift of the sigmoid along the x-axis:

$$exp(\varepsilon x - \delta) = exp(\varepsilon(x - \frac{\vartheta}{\varepsilon})) \qquad (A6)$$

For positive-defined $\varepsilon$ and $\vartheta$ $f(x) \in (0,1)$ and:

$$\begin{cases} f(x) \to 1, x < \frac{\vartheta}{\varepsilon}, \\ f(x) = 0.5, x = \frac{\vartheta}{\varepsilon}, \\ f(x) \to 0, x > \frac{\vartheta}{\varepsilon}. \end{cases} \qquad (A7)$$

Thus, the inflection point of the curve has the coordinates ($\vartheta/\varepsilon$; 0.5).

From (A2) and (A7), inflection point ($ip_{falling}$) of the falling edge has the coordinates:

$$ip_{falling}\left(\frac{e}{d}; F_0 + \frac{a}{2}\right) \tag{A8}$$

Similar to (A8), inflection point ($ip_{rising}$) of the rising edge has the coordinates:

$$ip_{rising}\left(\frac{c}{b}; F_0 + \frac{a}{2}\right) \tag{A9}$$

*A.3 Transition zone characteristics: accuracy*

We define start and end point of the transition zone as $x_1$ and $x_2$. Exponential function asymptotically tends to $F_0$ and $a$ values and cannot be calculated precisely. Thus, to obtain $x_1$ and $x_2$ we consider the level of accuracy to be equal to 1 HU. Since the equation contains dimensionless coefficients, the accuracy must be introduced into it being scaled by the signal's amplitude. Thus, taking into account typical expected $a$ values (100 - 500 HU [13](Greenway, K. et al., 2015)) we define accuracy as:

$$\delta y = \frac{1}{\max(a)} = \frac{1}{500} = 0.002 \tag{A10}$$

Therefore, the start and end points of the transition zone are defined as equal to baseline or plateau respectively, accounting the $\delta y * a$ correction:

$$\begin{cases} x_1: f(x_1) = baseline + \delta y * a, \\ x_2: f(x_2) = plateau - \delta y * a. \end{cases} \tag{A11}$$

*A.4 Transition zone characteristics: falling edge*

We demonstrate general approach for the positive sigmoid, corresponding to the falling edge. The start and end point coordinates correspond to the plateau and baselevel respectively(Figure A1b, green).

Start point ($x_1$) calculation:

$$F_0 + \frac{a}{1+exp(dx_1-e)} = F_0 + a - 0.002a, \tag{A12.1}$$

$$exp(dx_1 - e) = \frac{1}{0.998} - 1, \tag{A12.2}$$

$$dx_1 = \ln(0.002) + e, \tag{A12.3}$$

If we define $\ln(0.002)$ as a constant $\theta_1$, then (as $\ln(0.002) < 0$):

$$x_1 = \frac{1}{d}(e - \theta_1) \tag{A13}$$

End point ($x_2$) calculation:

$$F_0 + \frac{a}{1+exp(dx_2-e)} = F_0 + 0.002a, \tag{A14.1}$$

$$exp(dx_2 - e) = 500 - 1, \tag{A14.2}$$

$$dx_2 = \ln(499) + e, \tag{A14.3}$$

If we define $\ln(499)$ as a constant $\theta_2$, then (as $\ln(499) > 0$):

$$x_2 = \frac{1}{d}(e + \theta_2) \tag{A15}$$

Calculation of $\theta_1$ and $\theta_2$ shows equivalence of their absolute values:

$$\begin{cases} |\theta_1| = |\ln(0.002)| \approx 6.21, \\ |\theta_2| = |\ln(499)| \approx 6.21. \end{cases} \tag{A16}$$

Equivalence of $\theta_1$ and $\theta_2$ does not depend neither on the value of $a$ nor the accuracy, due to the properties of ln function:

$$|\theta_1| = \left|\ln\left(\frac{1}{1-\delta y} - 1\right)\right| = \left|\ln\left(\frac{1}{\delta y} - 1\right)\right| = |\theta_2| \tag{A17}$$

After transformation:

$$\left|\ln\left(\frac{1-\delta y}{\delta y}\right)\right| = |\ln(1-\delta y) - \ln(\delta y)| = |\ln(\delta y) - \ln(1-\delta y)| = \left|\ln\left(\frac{\delta y}{1-\delta y}\right)\right| \tag{A18}$$

Thus, we can define a constant $\theta$:

$$\theta = |\theta_1| = |\theta_2|, \tag{A19}$$

which value depends on chosen accuracy level and in general can be calculated as:

$$\theta = |\ln(\delta y)| \tag{A20}$$

The width of the transition zone ($\Delta x$) between plateau and baselevel is:

$$\Delta x_{falling} = |x_2 - x_1| = \left|\frac{1}{d}(e + \theta) - \frac{1}{d}(e - \theta)\right| = 2 * \frac{\theta}{d} \approx \frac{12.42}{d}, \tag{A21}$$

and depends only on $d$ coefficient.

*A.5 Transition zone characteristics: rising edge*

The start and end point coordinates correspond to the baselevel and plateau respectively (Figure A1b, light blue). To obtain the start and end points of the rising edge transition zone we need to solve two other equations:

$$\begin{cases} x_1: F_0 - \frac{a}{1+exp(bx_1-c)} = F_0 - a + a\delta y, \\ x_2: F_0 - \frac{a}{1+exp(bx_2-c)} = F_0 - a\delta y. \end{cases} \quad (A22)$$

The results of calculations will resemble (A21):

$$\Delta x_{rising} = |x_2 - x_1| = \left|\frac{1}{b}(c+\theta) - \frac{1}{b}(c-\theta)\right| = 2 * \frac{\theta}{b} \approx \frac{12.42}{b}, \quad (A23)$$

and depend only on $b$ coefficient.

*Edge slope*

We can take the first derivative of edge function (Figure A2) to determine the points through which the tangent will pass.

Rising edge:

$$\frac{df(x)}{dx} = \frac{a*b*exp(bx-c)}{(1+exp(bx-c))^2} \quad (A24)$$

Falling edge:

$$\frac{df(x)}{dx} = \frac{-a*d*exp(dx-e)}{(1+exp(dx-e))^2} \quad (A25)$$

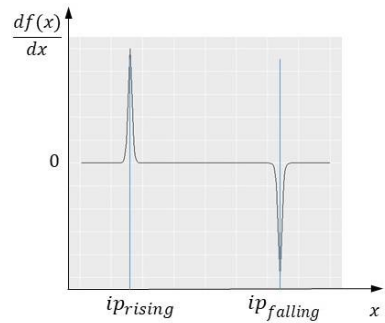

Figure A2 First derivative of the model curve

To define the slope angle, we need to calculate the first derivative at the inflection point.

Rising edge:

$$tg(\alpha) = \frac{df}{dx}\bigg|_{x=ip_{rising}} = \frac{a*b}{4} \quad (A26)$$

Falling edge:

$$tg(\beta) = \frac{df}{dx}\bigg|_{x=ip_{falling}} = -\frac{a*d}{4} \quad (A27)$$